\documentclass[11pt,a4paper]{article}

\usepackage[table]{xcolor}

\usepackage{times,latexsym}
\usepackage{url}
\usepackage[T1]{fontenc}

\usepackage[acceptedWithA]{tacl2018v2}

\usepackage{xspace,mfirstuc,tabulary}

\usepackage{microtype}
\usepackage{algorithm}
\usepackage{algorithmic}

\usepackage{amsmath}
\usepackage{amssymb}

\usepackage{booktabs}
\usepackage{bbding}
\usepackage{makecell}
\usepackage{multirow}
\usepackage{tabularx}
\usepackage{graphicx} 

\newif\iftaclinstructions
\taclinstructionsfalse 
\iftaclinstructions
\renewcommand{\confidential}{}
\renewcommand{\anonsubtext}{(No author info supplied here, for consistency with
TACL-submission anonymization requirements)}
\newcommand{\instr}
\fi

\iftaclpubformat 

\else

\fi

\title{OPAL: Ontology-Aware Pretrained Language Model for End-to-End Task-Oriented Dialogue}

\author{
 Zhi Chen$^1$, Yuncong Liu$^1$, Lu Chen$^1$\Thanks{The corresponding authors are Lu Chen and Kai Yu.}, Su Zhu$^2$, Mengyue Wu$^1$, Kai Yu$^1$\footnotemark[1] \\
 $^1$X-LANCE Lab, Department of Computer Science and Engineering \\ 
        MoE Key Lab of Artificial Intelligence, AI Institute, Shanghai Jiao Tong University\\
    State Key Lab of Media Convergence Production Technology and Systems, Beijing, China\\
    $^2$AISpeech Co., Ltd., Suzhou, China \\
 \texttt{\{zhenchi713, chenlusz, kai.yu\}@sjtu.edu.cn} \\
}

\date{}

\begin{document}
\maketitle
\begin{abstract}
This paper presents an ontology-aware pretrained language model (OPAL) for end-to-end task-oriented dialogue (TOD). Unlike chit-chat dialogue models, task-oriented dialogue models fulfill at least two task-specific modules: dialogue state tracker (DST) and response generator (RG). The dialogue state consists of the domain-slot-value triples, which are regarded as the user's constraints to search the domain-related databases. The large-scale task-oriented dialogue data with the annotated structured dialogue state usually are inaccessible. It prevents the development of the pretrained language model for the task-oriented dialogue. We propose a simple yet effective pretraining method to alleviate this problem, which consists of two pretraining phases. The first phase is to pretrain on large-scale contextual text data, where the structured information of the text is extracted by the information extracting tool. To bridge the gap between the pretraining method and downstream tasks, we design two pretraining tasks: ontology-like triple recovery and next-text generation, which simulates the DST and RG, respectively. The second phase is to fine-tune the pretrained model on the TOD data. The experimental results show that our proposed method achieves an exciting boost and get competitive performance even without any TOD data on CamRest676 and MultiWOZ benchmarks.
\end{abstract}

\section{Introduction}
A task-oriented dialogue system aims to assist users in accomplishing a specific task by interacting with natural language, i.e., reserving a hotel or booking flight tickets. With the popularity of the industrial dialogue system, the task-oriented dialogue system attracts extensive attention in research.

\begin{figure}[t]
\centering
\includegraphics[width=\columnwidth]{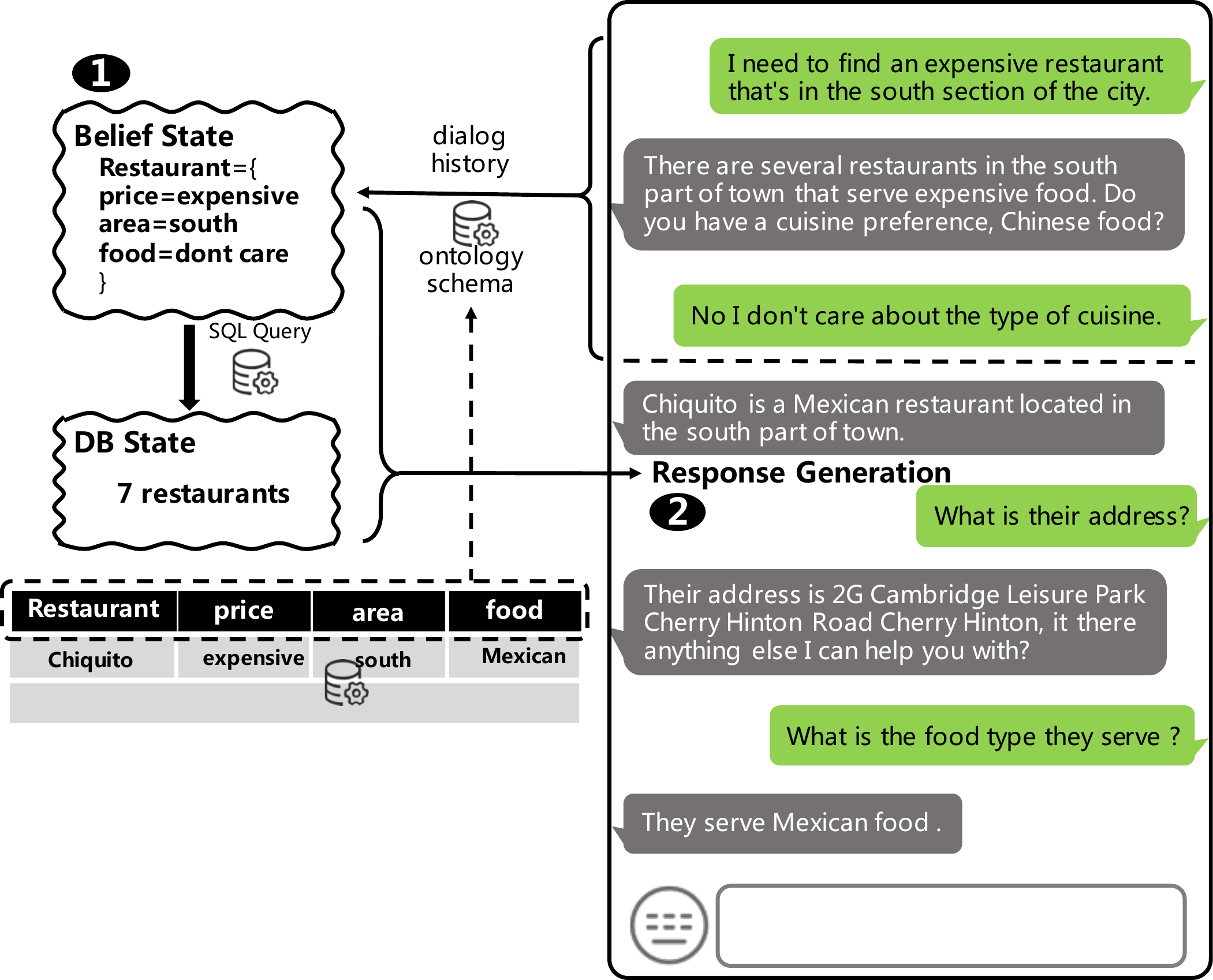} 
\caption{A task-oriented dialogue example. The dialogue model needs to infer the dialogue state based on the dialogue history and ontology schema. The DB state is searched by the generated dialogue state. The last step is to generate system response.}
\label{fig1}
\end{figure}

The existing task-oriented dialogue system can be classified into two categories: pipeline format and end-to-end format. The pipeline TOD system~\cite{ultes2017pydial,weisz2018sample} is composed of four modules: natural language understanding (NLU)~\cite{quirk2015language}, dialogue state tracking (DST)~\cite{xu2020memory,chen2020credit}, dialogue policy (DP)~\cite{chen2018policy,chen2019agentgraph,chen2020distributed} and natural language generation (NLG)~\cite{wen2015semantically,li2016deep,zhao2017learning}. Since each module of the system is trained separately and executes sequentially, it faces two serious issues: error accumulation and high annotation cost. Thus, the end-to-end dialogue system~\cite{lee2019convlab,zhao2019rethinking} gradually becomes the research focus, which formulates the task-oriented dialogue as a sequence-to-sequence task. The dialogue state, database (DB) state and the corresponding system response are directly concatenated together and flattened as a token sequence. The DB state is the status of the domain-related database searched with the dialogue state, as shown in Figure~\ref{fig1}.

Thanks to the success of pretraining language models~\cite{kenton2019bert,2020t5}, effective application has shed light on open-domain (chit-chat) dialogues~\cite{bao2020plato,adiwardana2020towards}. Nevertheless, utilizing such pretrained language models on TOD systems remains challenging due to the limited TOD data with annotated dialogue state. Unlike the open-domain dialogue, TOD is restricted by a dialogue ontology, which defines the dialogue domains, the slots and their candidate values. The TOD system needs to predict the dialogue state and feedback the DB content to accomplish a task. The dialogue state is structured information extracted from the dialogue context, which is a set of domain-slot-value triples. 

Recently, some works~\cite{hosseini2020simple,lin2020mintl} try to directly leverage the pretrained language models, e.g., GPT-2~\cite{radford2019language} and BART~\cite{lewis2020bart}, in the end-to-end TOD system. Such models~\cite{mehri2019pretraining} are pretrained on the large-scale contextual text with the general self-supervised method, e.g., language modeling and language denoising. However, in the task-oriented dialogue task, the dialogue state is structured information rather than a contextual text. The inconsistency between the pretrained and downstream tasks will impact the performance of the PLMs on the TOD benchmarks. To alleviate this problem, SOLOIST~\cite{peng2020soloist} fine-tunes the pretrained GPT-2 with the existing annotated TOD data and then transfers it to the other task-oriented dialogue generation tasks. Similarly, NCM~\cite{liu2021pretraining} first warm-ups the Transformer-based model with large-scale Reddit\footnote{\url{http://files.pushshift.io/reddit/}}~\cite{volske2017tl} data and then fine-tunes the model on the TOD data. However, the existing TOD data is too limited to pretrain a large-scale language model.

To alleviate the problems above and advance pretrained language model research, especially its application on TOD, we propose an \textbf{O}ntology-aware \textbf{P}retr\textbf{A}ined \textbf{L}anguage model (\textbf{OPAL}). From the high-level perspective, we can abstract the end-to-end TOD task into two sub-tasks: \textbf{ontology-like triple recovery} and \textbf{next-text generation}, which corresponds to dialogue state tracking task and response generating task. The ontology-like triple recovery in the TOD means to predict the corresponding value given the domain and the slot. The next-text generation is easy to design for the contextual text, which directly fulfills with masking the last sentence. The challenge is how to design the ontology-like triple recovery task, which needs to obtain the structured information from the contextual text. In this paper, we utilize the external OpenIE tools~\cite{angeli2015leveraging,kolluru2020openie6}\footnote{\url{https://github.com/dair-iitd/openie6}} to extract the relation triples (subject-relation-object) from the contextual text as the structured information. In most cases, the domain-slot-value triple can be regarded as relation triple, e.g., \textit{train-arrive-12:30}. The relation triples extracted from the contextual text can be regarded as the ontology-like triples. We design self-supervised \textbf{ontology-like triple recovery} task and \textbf{next-text generation} task to pretrain the model. 

The main contributions of this paper are summarized as below:
\begin{itemize}
    \item We leverage the external tool OpenIE to generate large amounts of TOD-like data, which is important for the development of pre-trained language models in the TOD community.
    \item To the best of our knowledge, this is the first work to design  \emph{self-supervised} tasks for end-to-end TOD tasks. It bridges the gap between pretrained language models and end-to-end TOD models.
    \item The experimental results show that our proposed pretrained model OPAL can get competitive performance even without any annotated TOD data in the pretraining process.
    \item Further fine-tuned on the annotated TOD data, our proposed method gets exciting performance gain on CamRest676 and MultiWOZ datasets.
\end{itemize}

\begin{figure*}[t]
\centering
\includegraphics[width=\textwidth]{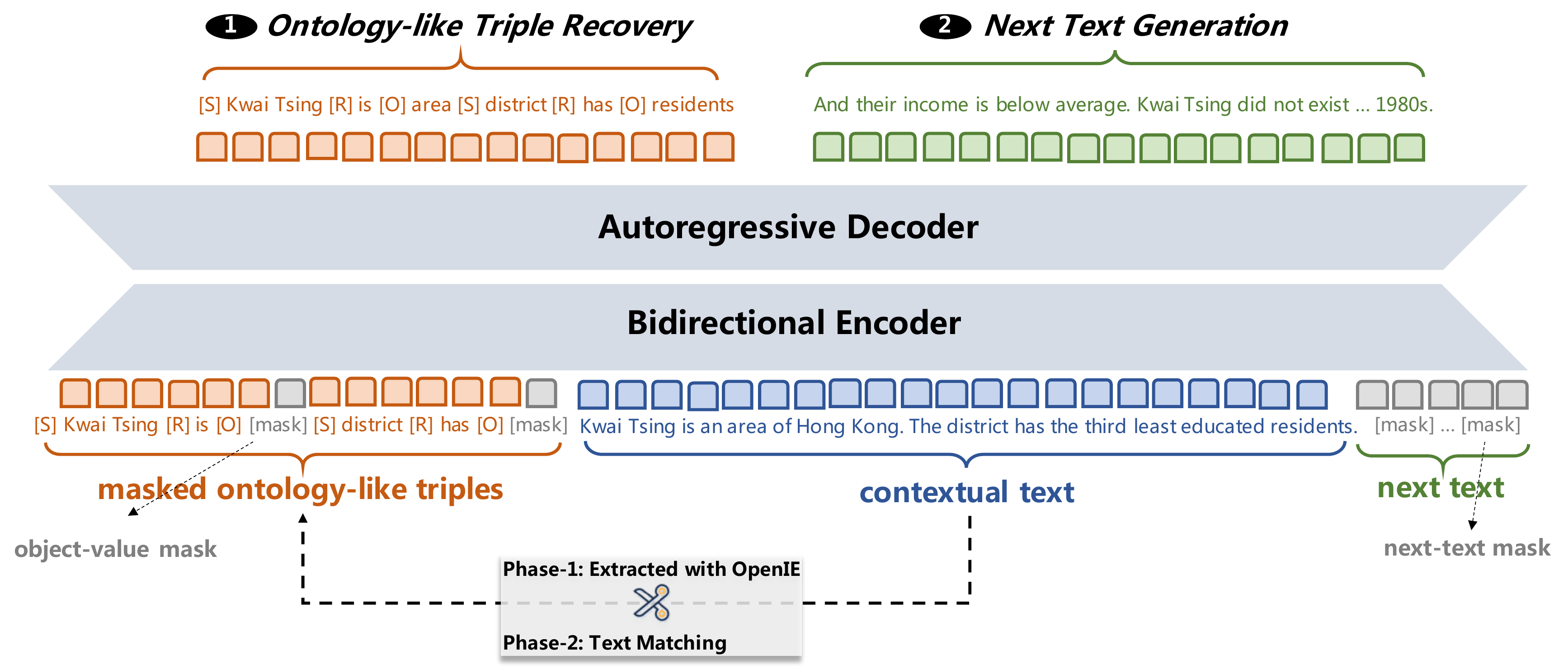} 
\caption{The ontology-aware pretraining method contains two masking strategies: object-value mask and next-text mask. The corresponding self-supervised learning methods are ontology-like triple recovery and next-text generation. The ontology-like triples of the contextual text are extracted by the external tool OpenIE at the pretraining phase-1 and matched with the given whole ontology at phase-2.}
\label{fig2}
\end{figure*}

\section{End-to-End Task-Oriented Dialogue}
As previously introduced, the pipeline dialogue system consists of four modules. The NLU module is to recognize the user's intents and the corresponding slot values. The DST module combines the previous state and the results of the NLU to update the current dialogue state. The DP module chooses the discrete dialogue acts according to the dialogue state and the database state to respond to the user. The NLG module is to generate the natural language based on the chosen dialogue acts. There are at least four kinds of annotation in such systems: the user's intent, the slot value, the dialogue state, and the dialogue act. The heavy annotation labor enormously increases the cost of building a pipeline system. Its poor scalability further influences the pipeline dialogue system development.

Compared with the pipeline system, this paper's end-to-end task-oriented dialogue system only requires the annotated dialogue state. The end-to-end TOD system is fed with the dialogue context $\mathbf{c}$ and generates the dialogue state $\mathbf{b}$ and delexicalized response $\mathbf{r}$, where the database (DB) state $\mathbf{d}$ is retrieved from the results searched with $\mathbf{b}$.
The delexicalized response means that the specific slot values are replaced with the corresponding slot placeholders. The lexicalized response is recovered from the delexicalized one with the generated dialogue state and DB state. The training sample at each dialogue turn of the end-to-end TOD model is defined as:
\begin{equation}
    \mathbf{x} = (\mathbf{c},\mathbf{b},\mathbf{d},\mathbf{r}).
\end{equation}
For the task-oriented dialogue, the dialogue context not only consists of the dialogue history $\mathbf{h}$ but also includes the dialogue ontology schema $\mathbf{s}$, which is usually ignored by the existing end-to-end models. The ontology can be seen as prior knowledge designed by the dialogue expert, which defines the dialogue domain, slots, and candidate values. The end-to-end TOD model needs to fulfill two sub-tasks: dialogue state tracking (DST) and response generation (RG). Formally, the learning goal of the TOD model is to maximize the joint probability $p_{\theta}(\mathbf{x})$, which can be factorized in an auto-regressive manner as:
\begin{align}
    p_{\theta}(\mathbf{x}) &= p(\mathbf{c},\mathbf{b},\mathbf{d},\mathbf{r}), \\
    &= p(\mathbf{h},\mathbf{s},\mathbf{b},\mathbf{d},\mathbf{r}), \label{eq:eq1}\\
    &= \underbrace{p(\mathbf{r}|\mathbf{b}, \mathbf{d}, \mathbf{h},\mathbf{s})}_{\rm RG} \underbrace{p(\mathbf{b}|\mathbf{h},\mathbf{s})}_{\rm DST} p(\mathbf{h},\mathbf{s}), \label{eq:eq2}
\end{align}
where the factorization from (\ref{eq:eq1}) to (\ref{eq:eq2}) is based on the fact that the database-lookup operation is a deterministic process. The $p(\mathbf{h},\mathbf{s})$ is the prior probability of the paired dialogue and ontology (as the input of the model), which depends on the distribution of the (pre-)training data and is independent on the model. The dialogue state tracker intrinsically extracts the ontology-related constraints demanded by the user, where the ontology schema is given in advance. 

\section{Ontology-Aware Pretraining Method}
The existing task-oriented dialogue data with the given ontology is limited to pretrain the language model. To increase the scale of the pretraining data, we divide the pretraining process into two phases. The first phase pretrains the model on the large-scale contextual text. The triples of the text are extracted by the latest neural-based OpenIE6~\cite{kolluru2020openie6}. There is still a glaring discrepancy between the contextual text and the dialogue. For example, the dialogue always contains co-reference and information ellipsis~\cite{iyyer2017search}. We pretrain the model on the smaller TOD data at the second phase to further decrease the gap between the pretrained model and the downstream tasks. The two phases are complementary to each other introduced as below:

\paragraph{Phase-1: Pretrained on Contextual Text} In traditional dialogue pre-trained models~\cite{zhang2020dialogpt}, the crawled Reddit data is popular to be used as pre-trained corpus. However, Reddit data contain lots of the co-reference and information ellipsis, which seriously impact the performance of the external information extraction tool. 
Different from the dialogue data, the co-reference and information ellipsis is infrequent in the contextual text of the Wikipedia\footnote{\url{https://dumps.wikimedia.org/enwiki/latest/enwiki-latest-pages-articles.xml.bz2}}. The more details are shown in Section~\ref{sec:abl} to validate the effects of pre-trained corpora.
We use the neural-based OpenIE6 to extract the ontology-like knowledge of contextual text automatically. We directly simulate the extracted subject-relation-object triples as the domain-slot-value triples. As shown in Figure~\ref{fig2}, the object values in the extracted ontology are masked during the pretraining process. One of our designed pretraining tasks is to recover the ontology-like triples (named \textbf{ontology-like triple recovery} as OR), which is similar to the DST task. To increase the inference ability of the pretrained model, we mask the next text (one or two sentences, which are randomly chosen.) and push the model to infer the next text (named \textbf{next-text generation} as NTG), which is similar to the RG task. Thus, the pretraining sample is composed of four elements: masked ontology-like triples $\hat{\mathbf{s}}$, the masked document context $\hat{\mathbf{h}}$, ontology-like triples $\hat{\mathbf{b}}$ and the next text $\hat{\mathbf{r}}$. Similar to Equation~\ref{eq:eq2}, the goal of the pretaining model is to maximize the joint probability:
\begin{align}
    p(\hat{\mathbf{h}},\hat{\mathbf{s}},\hat{\mathbf{b}},\hat{\mathbf{r}})= \underbrace{p(\hat{\mathbf{r}}|\hat{\mathbf{b}}, \hat{\mathbf{h}},\hat{\mathbf{s}})}_{\rm NTG} \underbrace{p(\hat{\mathbf{b}}|\hat{\mathbf{h}},\hat{\mathbf{s}})}_{\rm OR} p(\hat{\mathbf{h}},\hat{\mathbf{s}}).
\end{align}
To obtain the qualified triples of a sentence using OpenIE6, we remove all the stopwords in the triples and filter the triples that one of the triple components is a blank space. It is also the main reason that we do not choose the Reddit at this pretraining phase. There are lots of pronouns in the text, which is hard to extract qualified triples.
This pretraining phase vastly increases the scale of the pretraining data.
\begin{figure}[t]
\centering
\includegraphics[width=\columnwidth]{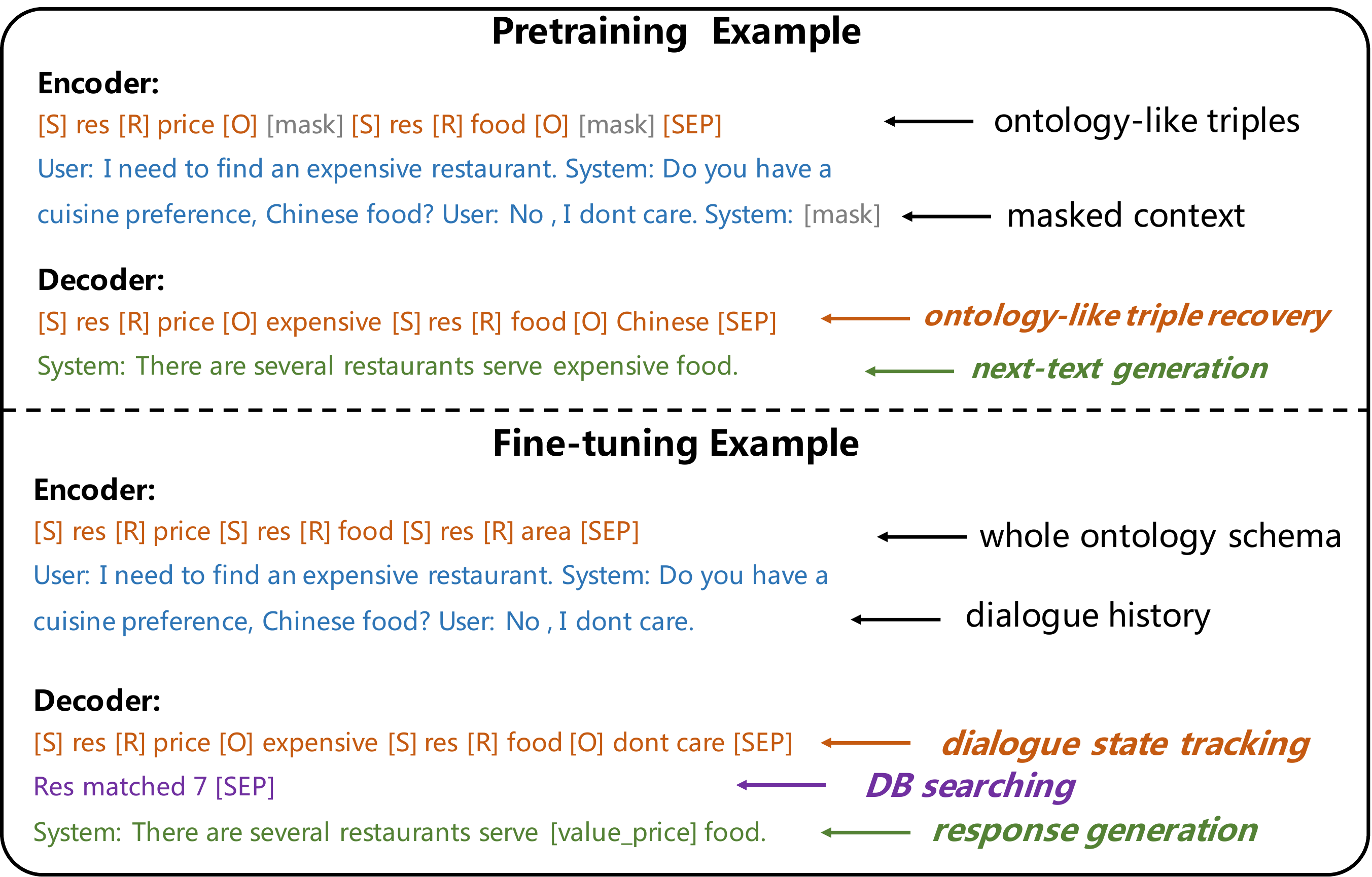} 
\caption{A toy example to show the differences between the pretraining data and the fine-tuning data.}
\label{fig3}
\end{figure}
There are four steps to filter the triples of the sentence:
\begin{itemize}
    \item Remove all the stopwords in the triples and filter the triples that one of triple component is a blank space.
    \item Remove the triples that one of triple component contains more than 4 words.
    \item For the triples that have the same subject-relation pair, randomly select one of the triples and remove the others.
    \item Randomly select two triples from the rest of triples, if its length is larger than two. This is to extract no more than two triples in a sentence.
\end{itemize}

\paragraph{Phase-2: Pretrained on TOD Data} To further decrease the gap between the pretrained language model and the end-to-end model, we leverage the smaller task-oriented data in the pretraining process. 
Instead of extracting the ontology-like triples with OpenIE6, the TOD ontology is designed by the dialogue experts. We directly use the text matching method to extract the domain-slot-value triples from the dialogue context with the given ontology. Note that the extracted triples with text matching operation are not the dialogue state. In this pretraining phase, the system-mentioned ontology triples also have to be recovered, which is consistent with the previous pretraining process. In other words, different from SOLOIST~\cite{peng2020soloist} and NCM~\cite{liu2021pretraining}, we do not need to use the annotated dialogue state and only utilize the given dialogue ontology to match the ontology-related triples. This attribute increases the generalization of the proposed ontology-aware pretraining methods, where the ontology is much easier to be obtained than the dialogue state annotation.
We give a toy example to distinguish the usage of the pretraining TOD data and the fine-tuning data of the end-to-end TOD task that shows in Figure~\ref{fig3}. At the pretraining process, the ontology is extracted from the context, which is just a part of the given ontology. The ontology recovery is to recover all the ontology-related triples, e.g., the triple res-food-Chinese is not in the dialogue state. At fine-tuning process, there is an extra database searching step.

\section{Experiments}
We evaluate our proposed pretrained model OPAL on dialogue state tracking tasks and end-to-end TOD tasks. To further validate the effectiveness of the proposed OPAL, we conduct the ablation study to analyze the effects of the different pretraining ingredients. Last but not least, we design the resource-limited experiments to figure out the sample efficiency of the proposed OPAL on the end-to-end TOD task and show some cases to study the strength of the proposed OPAL.

\subsection{Corpora}
At phase-1 of the proposed OPAL, we use the Wikipedia corpus to pretrain the model. There are 72.24 million samples collected from Wikipedia. We have used five task-oriented dialogue datasets in the experiments as shown in Table~\ref{tab:dataset}, where the Schema~\cite{rastogi2020towards} and the TaskMaster~\cite{byrne2019taskmaster} are leveraged in the phase-2 of the pretraining process and the rests are the downstream benchmarks. The WOZ~\cite{mrkvsic2017neural} and the CamRest676~\cite{wen2016conditional} are the single-domain task-oriented dialogue corpora, which are the well-studied DST benchmark and end-to-end TOD benchmark, respectively. The MultiWOZ is a kind of multi-domain dialogue corpus, which is challenging due to its multi-domain setting and diverse language styles. There are two versions of the MultiWOZ dataset used in the experiments: MultiWOZ2.0~\cite{budzianowski2018large} and MultiWOZ2.1~\cite{eric2019multiwoz}, where MultiWOZ2.1 fixes most of DST annotation errors in MultiWOZ2.0. 
To fairly compare to the other baselines, we run the end-to-end TOD tasks on the MultiWOZ2.0 and run the DST tasks on the MultiWOZ2.0 and MultiWOZ2.1.

\begin{table}[t]
\centering
\small
\newcolumntype{s}{>{\columncolor[HTML]{DCDCDC}} c}
\setlength{\tabcolsep}{0.02mm}{
\small
\begin{tabular}{c c c c c s} 
 \bottomrule
 \textbf{Dataset} & \textbf{\#Dialogue} & \textbf{\#Domain} & \textbf{\#Slot} & \textbf{X-Domain} & \textbf{Usage} \\ 
 \hline
 Schema & 22,825 & 17 & 123 & \Checkmark & P \\
 TaskMaster & 17,304 & 7 & 281 & \XSolidBrush & P \\
 MultiWOZ & 10,438 & 7 & 46 & \Checkmark & F \\
 WOZ & 1,200 & 1 & 4 & \XSolidBrush & F \\
 CamRest676 & 676 & 1 & 4 & \XSolidBrush & F \\
 \hline
\end{tabular}
}
\caption{The five task-oriented dialogue datasets are used in this paper. The X-domain (cross-domain) means that a dialogue can contain different dialogue domains. The usages of the datasets are grouped into Pretraining (named as P) and Fine-tuning (named as F), which mean that the corresponding dataset is used in pretraining phase and fine-tuning phase.}
\label{tab:dataset}
\end{table}

\subsection{Metrics}
For the dialogue state tracking task, we use the joint goal accuracy (\textbf{\textit{JGA}}) to evaluate the models. Only if all the predicted slot values at each turn are exactly matched with the golden, it says the successful prediction of the DST model. For the end-to-end TOD task, there are three reported scores: \textbf{\textit{Inform}}, \textbf{\textit{Success}} and \textbf{\textit{BLEU}}. The \textbf{\textit{Inform}} measures whether the system response has provided the right entity. The \textbf{\textit{Success}} reports whether the system response has 
provided all the requested slots. The \textbf{\textit{BLEU}} evaluates the naturalness of the generated system response. Following~\citet{budzianowski2018large}, the combined score (\textbf{\textit{Combined}}) is also reported using ${\rm \textbf{\textit{Combined}}} = ({\rm \textbf{\textit{Inform}}} + {\rm \textbf{\textit{Success}}}) \times {\rm 0.5} + {\rm \textbf{\textit{BLEU}}}$.

\begin{table*}[t]
\centering
\small
\newcolumntype{s}{>{\columncolor[HTML]{DCDCDC}} c}
\setlength{\tabcolsep}{1.5mm}{
\begin{tabular}{c c c c c c c} 
 \bottomrule
 \textbf{Model} & \textbf{Model Size} & \textbf{Dialogue Act} & \textbf{\textit{Inform}} & \textbf{\textit{Success}} & \textbf{\textit{BLEU}} & \textbf{\textit{Combined}} \\ 
 \hline
 Sequicity~\cite{lei2018sequicity} & - & \XSolidBrush & 66.40 & 45.30 & 15.54 & 71.39 \\
 HRED-TS~\cite{peng2019teacher} & - & \Checkmark & 70.00 & 58.00 & 17.50 & 81.50 \\
 DSTC8 Winner~\cite{ham2020end} & 124M & \Checkmark & 73.00 & 62.40 & 16.00 & 83.50 \\
 DAMD~\cite{zhang2020task} & - & \Checkmark & 76.40 & 60.40 & 16.60 & 85.00 \\
 SimpleTOD~\cite{hosseini2020simple} & 117M & \Checkmark & 84.40 & 70.10 & 15.01 & 92.26 \\
 SOLOIST~\cite{peng2020soloist} & 117M & \XSolidBrush & 85.50 & 72.90 & 16.54 & 95.74 \\
 MinTL-BART~\cite{lin2020mintl} & 406M & \XSolidBrush & 84.88 & 74.91 & 17.89 & 97.78 \\
 UBAR~\cite{yang2021ubar} & 82M & \XSolidBrush & 88.20 & 79.50 & 16.43 & 100.28 \\
 NCM$_B$~\cite{liu2021pretraining} & 116M & \Checkmark &  85.90 & 74.80 & 19.76 & 100.11 \\
 NCM$_L$~\cite{liu2021pretraining} & 292M & \Checkmark & 86.90 & 76.20 & 20.58 & 102.13 \\
 HTER~\cite{santra2021hierarchical} & - & \Checkmark & \textbf{91.72} & 75.80 & 19.05 & 102.81 \\
 \hline
 \hline
 BART & 139M & \XSolidBrush &  87.50 & 72.20 & 16.67 & 96.53 \\ 
 \rowcolor[HTML]{DCDCDC} OPAL & 139M & \XSolidBrush & 89.40 & 81.10 &	18.60 & \textbf{103.85} \\
 BART$_L$ & 406M & \XSolidBrush&  86.20 & 70.30 &	17.01 & 95.26 \\ 
 \rowcolor[HTML]{DCDCDC} OPAL$_L$ & 406M & \XSolidBrush & 88.00 & \textbf{82.80} & \textbf{20.80} & \textbf{106.20} \\
 \hline
\end{tabular}
}
\caption{End-to-end response generation results on MultiWOZ2.0. \Checkmark and \XSolidBrush denote whether the dialogue act annotation is used in the training process. List all the model sizes of the Transformer-based end-to-end TOD models. Notice that we directly use the UBAR result provided by ~\citeauthor{liu2021pretraining}. According the released code of the UBAR, they have not used the standard evaluation metric, which is unfair to compare to other methods. We also run their code with released model checkpoint, whose combined score is even worse than the result provided by ~\citeauthor{liu2021pretraining}. Results are significant (p < 0.01) comparing OPAL model and BART model as the initialized TOD model.}
\label{tab:multiwoz_e2e}
\end{table*}

\subsection{Experimental Setup}
We implement the proposed OPAL with HuggingFace’s Transformers~\cite{wolf-etal-2020-transformers} and BART, which is a pretrained denoising autoencoder. To validate the generalization of the proposed pretraining method, we set the base version and large version (BART$_L$) of the BART as the backbone of the proposed OPAL named OPAL and OPAL$_L$ respectively. The learning rates of the pretraining and fine-tuning are both 1e-5. The optimizer is AdamW. At phase-1 of the pretraining process, the total training step is 280,000 and the batch size is 256. It is pre-trained on four P100 GPUs (16G memory for each). This pretraining process costs 260 hours (one epoch on Wikipedia). Similar to NCM~\cite{liu2021pretraining}, we pretrain 100,000 steps at the phase-2. At the fine-tuning process of the downstream tasks, the batch size is 32. We conduct significant tests with five different seeds on the end-to-end TOD task, where the final results are trained with the default seed 42.

\subsection{Baselines}
We compare the proposed OPAL with the strong baselines, which hold the state-of-the-art (SOTA) performance on the DST and end-to-end TOD. 

The DST models can be divided into two categories: classification method and generation method. The classification methods rely on the optional slot values of the ontology and select the value from it. Their scalability is a severe problem for the practical dialogue system. The generation methods directly extract the values from the dialogue context, which are comparable to the proposed OPAL.

For the end-to-end TOD tasks, the existing end-to-end TOD systems can be grouped into modular systems and sequential systems. The modular systems use multiple decoders to generate the downstream outputs independently and are trained in an end-to-end manner.  The sequential systems formulate the end-to-end TOD as a single sequence prediction problem. Sequicity~\cite{lei2018sequicity} proposes a two-stage CopyNet method to generate the dialogue state and the system response. HRED-TS~\cite{peng2019teacher} proposes a teacher-student framework with a hierarchical recurrent encoder-decoder backbone. DAMD~\cite{zhang2020task} designs a domain-aware multi-decoder network with the multi-action data augmentation method. DSTC8 Winner~\cite{ham2020end} and SimpleTOD~\cite{hosseini2020simple} successfully leverage the pretrained language model GPT-2 for the end-to-end TOD modelling in the unified way. Inspired by SimpleTOD, SOLOIST~\cite{peng2020soloist} fine-tunes GPT-2 with out-of-domain TOD data and gets excellent transferability. MinTL-BART~\cite{lin2020mintl} and UBAR~\cite{yang2021ubar} improve the end-to-end TOD system by changing the input content without extra assumptions.
HTER~\cite{santra2021hierarchical} improves the end-to-end TOD system by hierarchical dialogue modeling mechanism.
NCM~\cite{liu2021pretraining} improves the decoder with the noisy channel model and proposes a two-stage pretrianing method to warm up the Transformer-based model, where the model first pretrains on the Reddit corpus and then on the task-oriented dialogues. NCM is the closest method with our proposed method. We mainly compare our proposed method with this method.

\begin{table}[t]
\centering
\setlength{\tabcolsep}{1.5mm}{
\small
\newcolumntype{s}{>{\columncolor[HTML]{DCDCDC}} c}
\begin{tabular}{c c c c c} 
 \bottomrule
 \textbf{Model} & \textbf{\textit{Inform}} & \textbf{\textit{Success}} & \textbf{\textit{BLEU}} & \textbf{\textit{Combined}} \\ 
 \hline
 Sequicity & 92.30 & 85.03 & 21.40 & 110.20 \\
 SOLOIST & 94.70 & 87.10 & 25.50 & 116.40 \\
 NCM$_B$ & 94.30 & 85.20 & 25.98 & 115.73 \\
 NCM$_L$ & 95.40 & 85.30 & \textbf{26.89} & 117.24 \\
 \hline
 \hline
 BART &  96.31 & 79.41 & 24.74 & 112.61 \\
 \rowcolor[HTML]{DCDCDC} OPAL & \textbf{96.32} & \textbf{89.86} & 26.56 & \textbf{119.65} \\
 \hline
\end{tabular}
}
\caption{End-to-end response generation results on CamRest676.}
\label{tab:camrest_e2e}
\end{table}

\subsection{Results on End-to-End TOD}
We first fine-tune our pretrained models OPAL and OPAL$_L$ on two well-studied end-to-end TOD datasets: MultiWOZ2.0 and CamRest676, as shown in Table~\ref{tab:multiwoz_e2e} and Table~\ref{tab:camrest_e2e}. We compare our models with strong baselines in the end-to-end dialogue learning setting. 

To validate the generalization of our proposed ontology-aware pretraining method, we set the base-version and large-version BART as the backbones of the pretraining models.
Compared with the performance fine-tuned on the original BARTs, the proposed OPAL and OPAL$_L$ achieve 7.32 and 10.94 overall performance gains on the MultiWOZ2.0 dataset and absolute 7.04 point gains on the CamRest676 dataset. SOLOIST~\cite{peng2020soloist} and NCM~\cite{liu2021pretraining} are the two closest methods to OPAL, which both leverage the out-of-domain TOD in pretraining the Transformer-based models. Different from our methods, these two approaches rely on DST annotation. Our proposed models can still get the best task completion (Inform and Success) and have lower BLEU scores than NCM barely. Compared with overall baselines, our proposed models reach the new SOTA overall performance (Combined) on both two datasets. The large-version model OPAL$_L$ outperforms the base-version OPAL with a 2.53 performance gain on the combined score. 
To fairly compare to other baselines, we only report the base-version OPAL's performance in the next experiments.

\begin{figure}[t]
\centering
\includegraphics[width=\columnwidth]{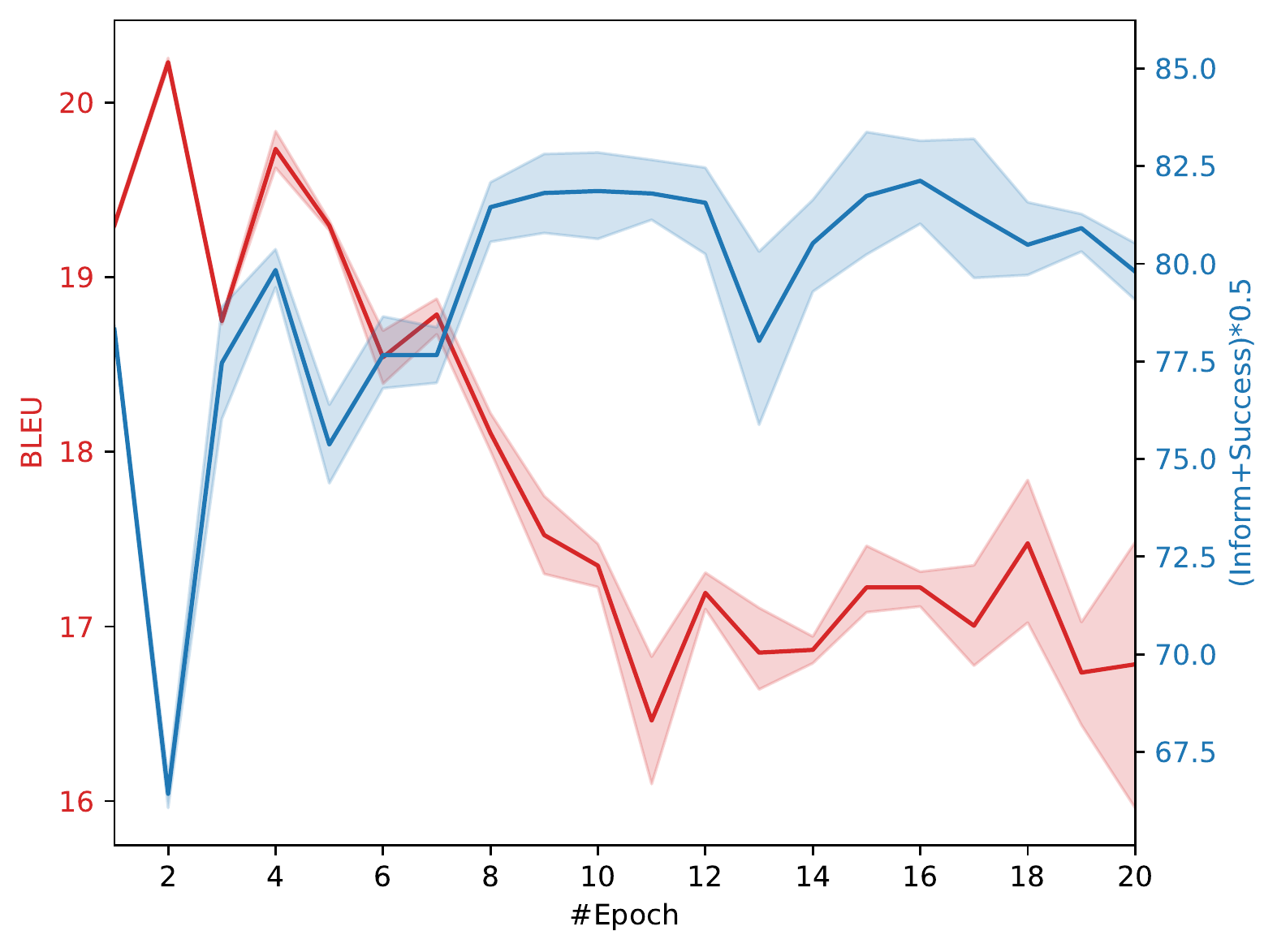}
\caption{The correlation between BLEU score and task-completion ability at first 20 finetuning epochs. They are the average evaluation results on MultiWOZ2.0 with different five seeds.}
\label{fig8}
\end{figure}

Compared with NCM$_B$, our proposed OPAL has higher task-completion (revealed by ${\rm \textbf{\textit{Inform}}} + {\rm \textbf{\textit{Success}}}) \times {\rm 0.5}$) performance. However, BLEU score of OPAL is lower than BLEU of NCM$_B$. Figure~\ref{fig8} shows the correlation between BLEU score and task-complation ability. The fine-tuned model tried to balance between BLEU score and task-completion ability. With the progress of training process, the BLEU score is descending and the task-completion ability is enhanced. The main reason is that there are different expressions on the same system intention, which is the typical one-to-many mapping problem~\cite{zhao2018zero} in the dialogue generation. The final fine-tuned model has stronger task-completion ability but sacrifices the dialogue diversity. In the evaluation, we choose the model with the highest combination score.

\begin{table}[t]
\centering
\small
\newcolumntype{s}{>{\columncolor[HTML]{DCDCDC}} c}
\setlength{\tabcolsep}{1.0mm}{
\begin{tabular}{c c c}
 \bottomrule
 \multirow{2}{6em}{\centering \textbf{Model}} & \multicolumn{2}{c}{\makecell[c]{\textbf{\textit{JGA}} \\ MultiWOZ}} \\ 
 \cline{2-3}
 & 2.0 & 2.1 \\
 \hline
 FJST~\cite{eric2017key} & 40.20 & 38.00 \\
 HyST~\cite{goel2019hyst} & 44.24 & - \\
 SUMBT~\cite{lee2019sumbt} & 46.65 & - \\
 TOD-BERT~\cite{wu2020tod} & - & 48.00 \\
 DST-Picklist~\cite{zhang2020find} & - & 53.30 \\
 SST~\cite{chen2020schema} & 51.17 & 55.23 \\
 TripPy~\cite{heck2020trippy} & - & 55.29 \\
 FPDSC~\cite{zhou2021dialogue} & \textbf{53.17} & \textbf{59.07} \\
 \hline
 TRADE~\cite{wu2019transferable} & 48.62 & 45.60 \\
 COMER~\cite{ren2019scalable} & 48.79 & - \\
 NADST~\cite{Le2020Non-Autoregressive} & 50.52 & 49.04 \\
 DSTQA~\cite{zhou2019multi} & 51.44 & 51.17 \\
 SOM-DST~\cite{kim2020efficient} & 51.38 & 52.57 \\
 MinTL-BART~\cite{lin2020mintl} & 52.10 & 53.62 \\
 SimpleTOD~\cite{hosseini2020simple} & - & 55.72 \\
 UBAR~\cite{yang2021ubar} & 52.59 & 56.20 \\
 SOLOIST~\cite{peng2020soloist} & 53.20 & 56.85 \\
 \rowcolor[HTML]{DCDCDC} OPAL & \textbf{54.10} & \textbf{57.05} \\
 \hline
\end{tabular}
}
\caption{Dialogue state tracking results on MultiWOZ2.0 and MultiWOZ2.1. The upper part is classification-based model and the lower part belongs to generation-based model.}
\label{tab:multiwoz_dst}
\end{table}

\subsection{Results on DST}
The classification-based DST models and generation-based DST models show at the upper part and lower part of the Table~\ref{tab:multiwoz_dst} and Table~\ref{tab:woz_dst} respectively. Table~\ref{tab:multiwoz_dst} reports the DST results on the MultiWOZ2.0 and MultiWOZ2.1 datasets. 

Our proposed OPAL can get the highest joint goal accuracy (JGA) among all the generation-based baselines on both datasets. 
Compared with the classification-based SOTA model FPDSC~\cite{zhou2021dialogue}, OPAL can even achieve 0.93\% JGA improvement on MultiWOZ2.0 dataset. Table~\ref{tab:woz_dst} shows the DST results on the WOZ, which is a single-domain dataset and has only 4 slots. The computational complexity of the classification-based models is proportional to the number of the candidate slot values. The classification-based models have the advantage of predicting slot values from valid candidates on the simpler dialogue domain. It is the main reason that the classification-based models are more popular on the single-domain WOZ dataset. Compared with the well-designed classification-based model BERT-DST~\cite{lai2020simple}, OPAL has a 0.7\% JGA gain. OPAL gets 6.7\% higher joint goal accuracy over the novel generation-based model TRADE~\cite{wu2019transferable}. Notice that we do not compare the proposed model with variants~\cite{yu2020score,li2020coco,dai2021preview} of the data augmentation methods based on TripPy~\cite{heck2020trippy}. In this paper, we pay more attention on the end-to-end task-oriented dialogue generation task. Our model is totally compatible with these data augmentation methods. In the future, we will try these augmentation methods on our model.

\begin{table}[t]
\centering
\small
\newcolumntype{s}{>{\columncolor[HTML]{DCDCDC}} c}
\begin{tabular}{c c} 
 \bottomrule
 \textbf{Model} & \makecell[c]{\textbf{\textit{JGA}} \\ WOZ} \\ 
 \hline
 NBT~\cite{mrkvsic2017neural} & 84.4 \\
 GLAD~\cite{zhong2018global} & 88.1 \\
 GCE~\cite{nouri2018gce} & 88.5 \\
 G-SAT~\cite{balaraman2019scalable} & 88.7 \\
 StateNet~\cite{ren2018towards} & 88.9  \\
 BERT-DST~\cite{lai2020simple} & 90.5 \\
 \hline
 TRADE~\cite{wu2019transferable}$^{\dag}$ & 84.5 \\
 \rowcolor[HTML]{DCDCDC} OPAL & \textbf{91.2} \\
 \hline
\end{tabular}
\caption{Dialogue state tracking results on the singel-domain WOZ. The upper part is classification-based model and the lower part belongs to generation-based model. ${\dag}$ represents the result is produced by us from their released code.}
\label{tab:woz_dst}
\end{table}

\begin{table}[h!]
\centering
\small
\newcolumntype{s}{>{\columncolor[HTML]{DCDCDC}} c}
\setlength{\tabcolsep}{0.8mm}{
\begin{tabular}{c | c c c c} 
 \bottomrule
  \hline
 \multirow{2}{4em}{\centering \textbf{Model}} & \multicolumn{4}{c}{MultiWOZ2.0} \\ 
 & \textbf{\textit{Inform}} & \textbf{\textit{Success}} & \textbf{\textit{BLEU}} & \textbf{\textit{Combined}} \\
 \hline
 OPAL & \textbf{89.40} & \textbf{81.10} & \textbf{18.60} & \textbf{103.85} \\
\hline
 \multicolumn{5}{l}{\textbf{Effect of Pre-trained Corpora}} \\
 WIKI &  88.40 & 79.50 & 18.28 & 102.23 \\
 TOD & 89.00 & 78.20 & 17.55 & 101.15 \\
 REDD &  86.90 & 77.10 & 16.93 & 98.93 \\
 \hline
 \multicolumn{5}{l}{\textbf{Effect of Pre-trained Tasks}} \\
 w/o NTG & 87.00 & 80.80 & 16.88 & 100.79 \\
 w/o OR & 85.20 & 79.50 & 17.52 & 99.88 \\
 \hline
 \multicolumn{5}{l}{\textbf{Effect of IE Tools}} \\
 OpenIE-Stanford & 88.40 & 79.20 & 17.34 & 101.14 \\
 \hline
  BART & 87.50 & 72.20 & 16.67 & 96.52 \\
 \hline
\end{tabular}
}
\caption{Ablation study on MultiWOZ2.0. There are three types of ablation study. The first is to analyze the effects of the pre-trained data. The second is to validate the effects of the designed pre-trained tasks. The last is to figure out the effects of IE tools. Results are significant (p < 0.01) comparing OPAL model and BART model as the initialized TOD model.}
\label{tab:ablation}
\end{table}

\section{Analysis}
The analysis experiments evaluate the proposed OPAL on the end-to-end TOD tasks to answer three main questions: \textbf{Q1:} What role do the different pretraining corpora (Wikipedia and out-of-domain TOD) play? \textbf{Q2:} What is the main factor that affects the pre-trained model? \textbf{Q3:} Does OPAL have a higher sample efficiency than the original BART in the limited-resource setting?

\begin{figure*}[t]
\centering
\includegraphics[width=\textwidth]{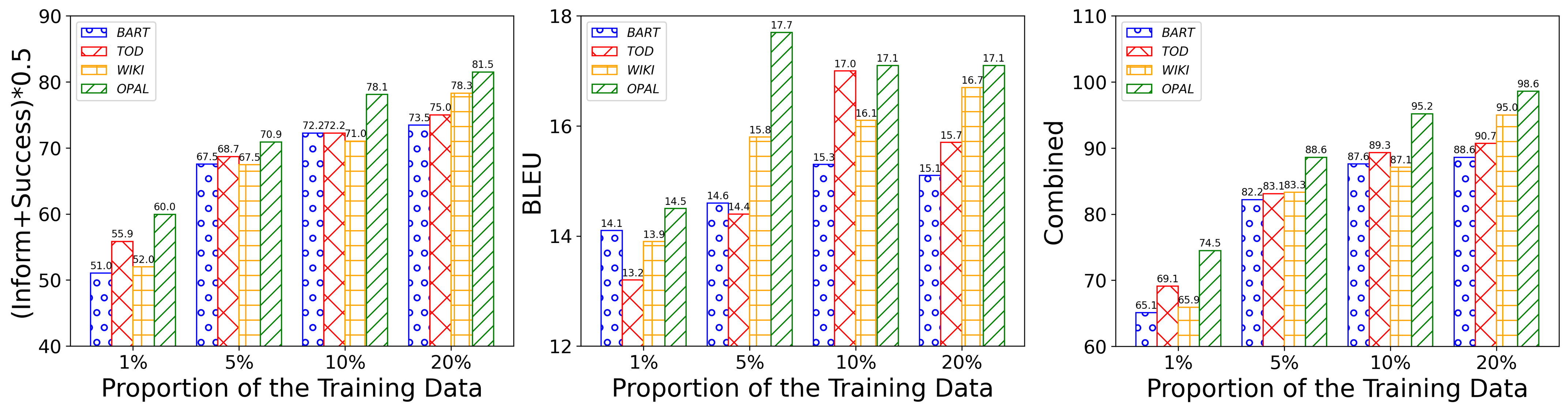} 
\caption{Resource-limited response generation results on MultiWOZ2.0. 1\% (80 dialogues), 5\% (400 dialogues), 10\% (800 dialogues) and 20\% (1600 dialogues) of training data are used to train each model.}
\label{fig4}
\end{figure*}

\begin{figure}[t]
\centering
\includegraphics[width=\columnwidth]{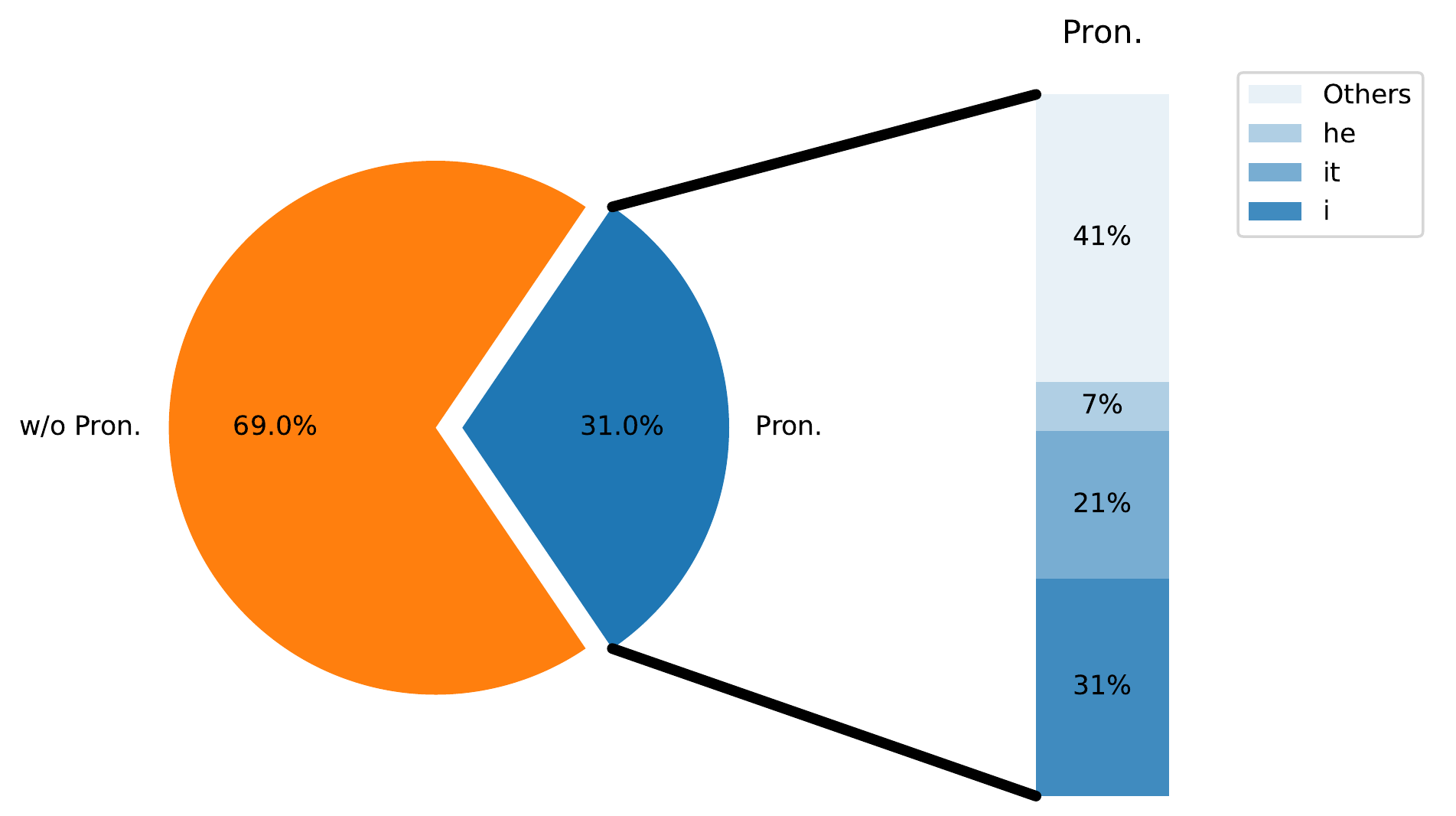}
\caption{The occupation rate of the extracted triples that contained pronouns as subject or object in Reddit corpus with OpenIE6.}
\label{fig9}
\end{figure}

\subsection{Ablation Study}
\label{sec:abl}
Table~\ref{tab:ablation} reports the ablation study of the proposed OPAL, which has two pretraining phases. The phase-1 of OPAL pretrains on the contextual texts and phase-2 pretrains on the task-oriented dialogues with the ontology-aware pretraining method. To evaluate the effects of these two corpora, we separately pretrain the backbones (BART) only on the pretraining data contextual texts or task-oriented dialogues, where the pretrained models are named as WIKI and TOD respectively. 
The pretrained models WIKI and TOD still outperform the original BART by a large margin. It indicates the efficiency of the proposed ontology-aware pretraining method. Especially, the pretrained model WIKI that does not see any TOD data at the pretraining phase can get the competitive performance with the NCM$_L$. Compared with NCM$_B$ with a similar parameter scale to our model, WIKI has apparent advantages on both end-to-end TOD datasets. The WIKI has the better performance with TOD. We know that WIKI suffers from the unseen TOD data and TOD suffers from the scale of the pretraining data. Our proposed OPAL adopts a two-stage pretraining method to solve the above problem, a classic example of ``one plus one greater than two''. The two-stage pretrained model OPAL outperforms the separated one with a 1.65 and 2.70 upper combined score on MultiWOZ2.0. It indicates that the ontology-aware contextual text corpus and ontology-aware TOD data are complementary. 

To further compare Wikipedia to Reddit corpus, we also use the same scale of Reddit data to conduct the Phase-1 pre-training, named REDD. WIKI is ahead of REDD in all the automatic metrics (BLEU and task-completion). To deeply analyze the effect factor, we calculate the occupation rate of the extracted triples that contained the pronouns as subject or object. As shown in Figure~\ref{fig9}, there are 31.0\% triples in Reddit data containing the pronouns. The highest frequency of pronouns is ``i'', which occupies 31\%. There are only 0.7\% triples contained pronouns in Wikipedia. In the TOD, the domains and slot values in the dialogue states are specific entities, which are not pronouns. The meaningless pronouns increase the gap between pre-training model and TOD model. The co-reference and information ellipsis in Reddit seriously hurt the performance of the external information extraction tool. It is the main reason that we choose the Wikipedia as the pre-training corpus.

We also evaluate the effects of the pre-trained tasks: ontology-like triple recovery (OR) and next-text generation (NTG). We directly remove the extracted triples in the input in ``w/o OR'' study. The ``w/o NTG'' means that the model only needs to recover the masked triples. The results show that OR task and NTG task benefit the task completion and the contextual consistency respectively. In the complex dialogue domain, the single-task pre-trained methods can not achieve comparable performance with OPAL. It indicates that the two designed tasks are both significant to reduce the gap between pre-trained model and TOD model.

We further validate the effects of different OpenIE tools. In our main experiments, we use the latest neural-based IE tool OpenIE6. There is also a very popular rule-based IE tool OpenIE-Stanford. Compared with OpenIE-Stanford, Neural-based OpenIE6 achieves promising performance improvement on well-studied IE benchmarks~\cite{kolluru2020openie6}. As shown in Table~\ref{tab:ablation}, WIKI with OpenIE6 is also better than OpenIE-Stanford tool in all the metrics. However, the improvement of neural-based OpenIE6 is limited, which indicates that the proposed pre-training method is not sensitive about IE accuracy.

\subsection{Sample Efficiency}
Under the different resource-limited settings, the proposed OPAL can get all the best performance in terms of task completion (Inform and Success), response naturalness (BLEU) and overall performance among the baselines, as shown in Figure~\ref{fig4}. It indicates the sample efficiency of the proposed ontology-aware pretraining method. When the training data is extremely limited (only 80 dialogues), TOD can improve overall performance by a large margin (absolute 3.2 point improvement) than WIKI. This improvement comes from the task completion ability, which indicates the TOD data can increase the generalization of the pretrained model for end-to-end TOD tasks. With the training data increase, WIKI pretrained on the large-scale context text data has the larger performance gain than TOD. When the number of the training data reaches 1600 dialogues, WIKI gets absolute 4.8 point gains over TOD. It indicates that the scale of the pretraining data influences the growth potential of the pretrained model. On the other hand, TOD outperforms over the WIKI in three of four data limitation cases on task-completion ability. However, WIKI achieves better performance on fluent statement (revealed by ${\rm \textbf{\textit{BLEU}}}$). It indicates that WIKI benefits the task-completion ability and TOD facilitates fluency and context consistency.

\begin{figure}[t]
\centering
\includegraphics[width=\columnwidth]{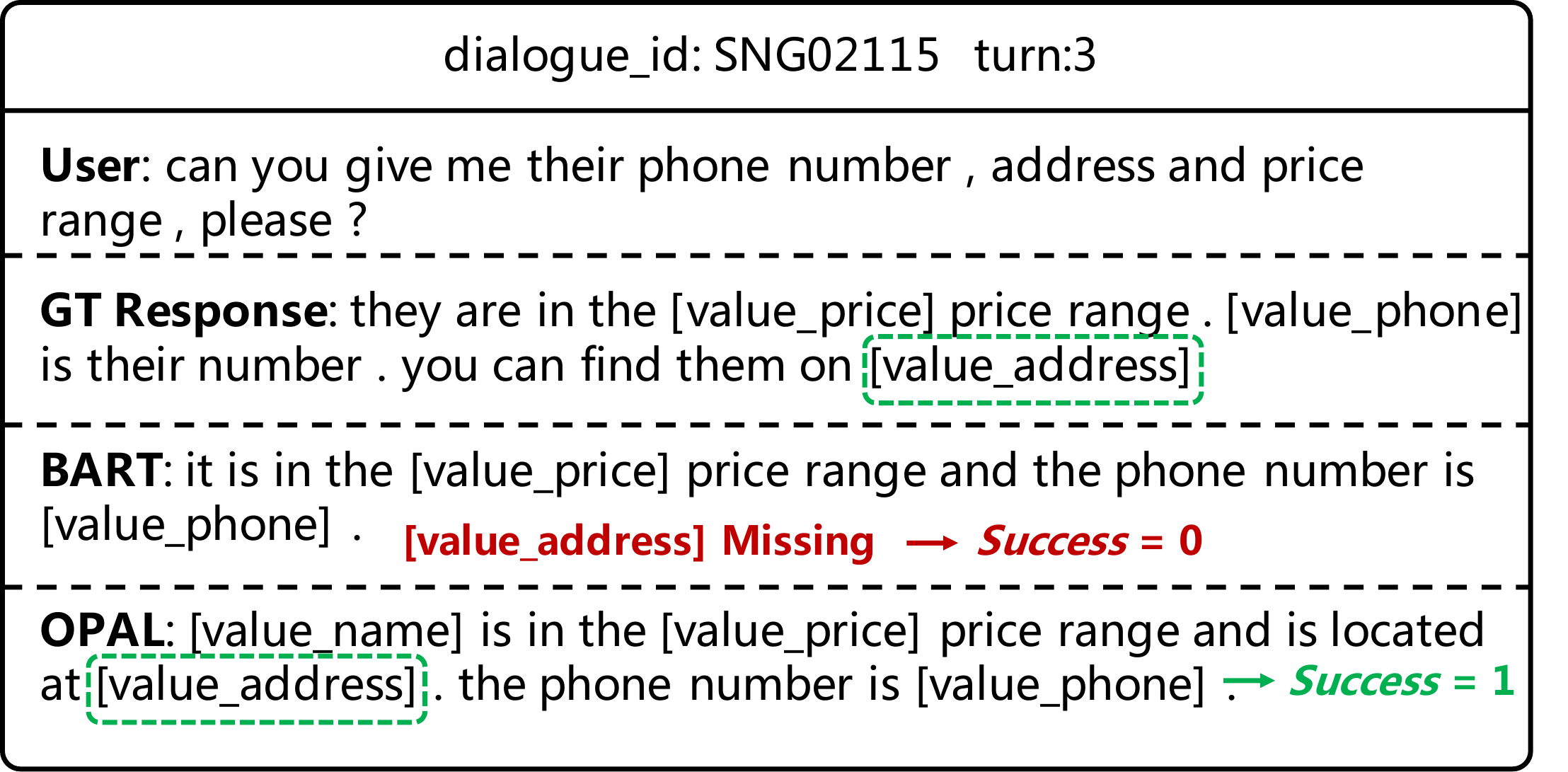}
\caption{Third dialogue turn in the dialogue session SNG02115 from MultiWOZ2.0 development set. The oracle response is represented as GT Response. BART and OPAL means that the responses are generated by the corresponding models.}
\label{fig5}
\end{figure}

\subsection{Case Study}
\begin{figure}[t]
\centering
\includegraphics[width=\columnwidth]{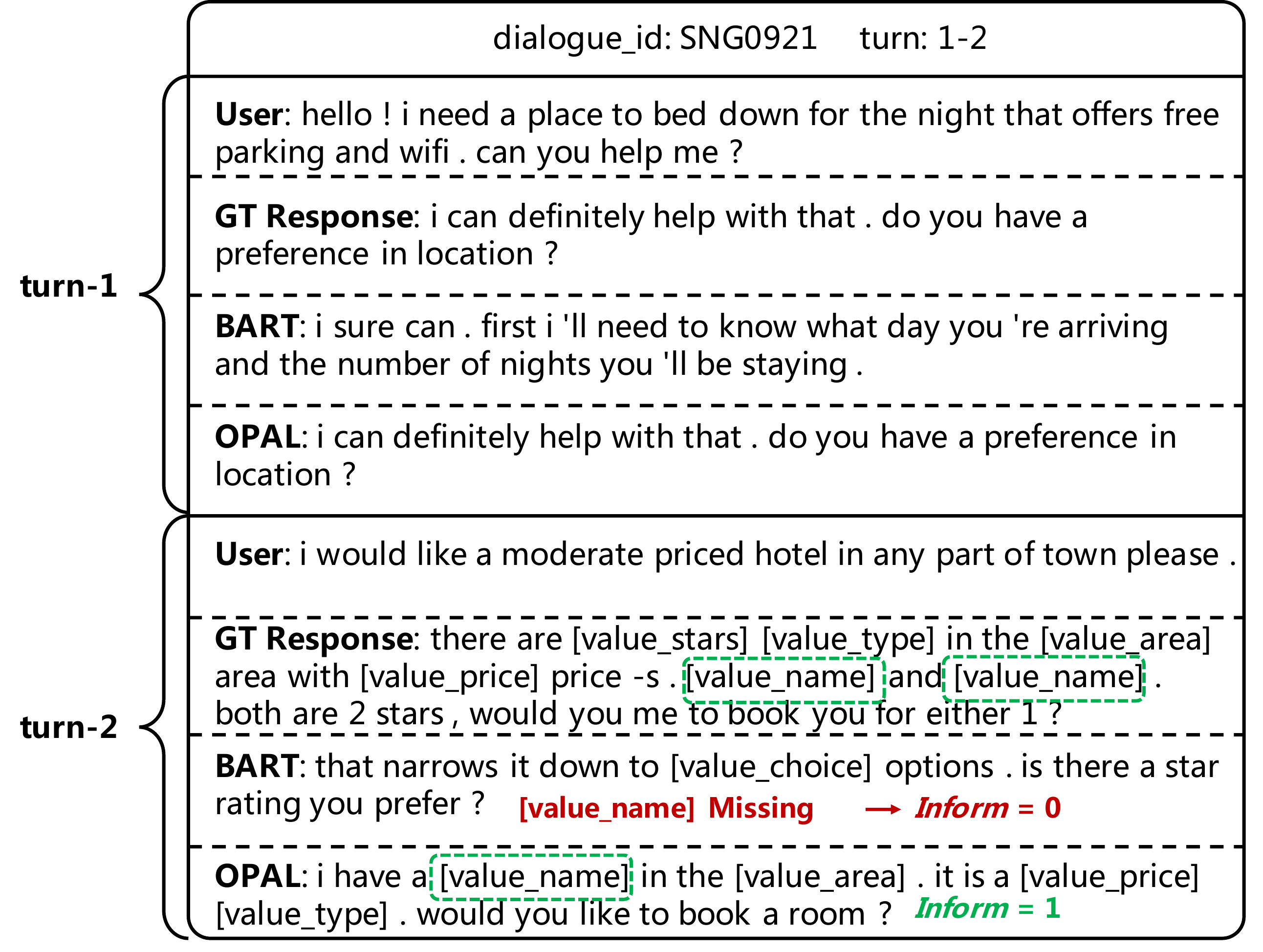}
\caption{The first two dialogue turns in the dialogue session SNG921 from MultiWOZ2.0 development set. The oracle response is represented as GT Response. BART and OPAL means that the responses are generated by the corresponding models.}
\label{fig6}
\end{figure}

Our proposed pretrained model OPAL has improved the performance on task completion and contextual consistency over the original BART. As shown in Figure~\ref{fig5}, we can see that the dialogue model fine-tuned from BART misses responding a request (address) to the user. Instead, our proposed OPAL accurately provides all the requested information to the user. As shown in Figure~\ref{fig6}, at the first turn, we can see that our proposed OPAL can provide the more similar response as the oracle than BART. It indicates that OPAL has the better performance on the response prediction. At the second turn, the dialogue system needs to provide the correct entity to the user. The original BART model chooses to miss it. Our proposed OPAL recommends an entity to the user in time. Compared with the original BART, the proposed OPAL has a obvious advantage in modeling the task-oriented dialogue, which not only generates the precise response but also completes the dialogue task successfully. This performance improvement comes from the two-stage ontology-aware pretraining method on the large-scale contextual text with the handcrafted ontology-like triples and the small task-oriented dialogue data with given ontology.


\section{Related Work}
\paragraph{End-to-End TOD Systems}
Early studies for end-to-end task-oriented dialogue systems either design a neural network-based model or propose a reinforcement learning method to use the reward signal to update the whole system. In these systems, the modules in the pipeline TOD system still exist and need their separated annotation. These systems usually can get promising performance on one specific task but have poor transferability. With the emergence of the multi-domain TOD benchmark, like MultiWOZ, the generative DST method has replaced the classification method as the mainstream over recent years due to its better generalization ability. It encourages formulating the end-to-end TOD as a text-to-text task. \citet{lei2018sequicity} propose a two-stage CopyNet to generate the dialogue state and response jointly with a single seq2seq architecture. \citet{zhang2020task} design a data augmentation method to increase the response diversity. The dialogue state, dialogue act and the response are generated with a shared encoder and the different decoders. Note that our proposed model does not use the annotated dialogue acts. Recently, some works \cite{hosseini2020simple,peng2020soloist,lin2020mintl,yang2021ubar} directly leverage the pretrained language models (like GPT-2 and BART) as the end-to-end TOD model in a unified way. \citet{liu2021pretraining} propose a Transformer-based noisy channel method to model the response prior and use the Reddit data and TOD data to warm up the TOD model. Most recently, \citet{su2021multi} formulate all the end-to-end TOD tasks as the unified generation tasks, which learns in multitask learning manner. \citet{he2021galaxy} propose a semi-supervised method to explicitly learn dialogue policy from limited labeled dialogues. Our proposed pre-trained method is compatible with these end-to-end TOD training strategies.

\paragraph{Self-supervised Learning for Dialogue System} Recent advances in supervised learning have witnessed the success of the pretrained language models on language understanding and generation tasks. Since the large-scale comment data in Reddit can be regarded as a kind of chit-chat dialogue, the self-supervised methods have been used in the chit-chat systems first. DialoGPT~\cite{zhang2020dialogpt} adapts the pretrained GPT-2 in the large-scale dialogue data. PLATO~\cite{bao2020plato} proposes a discrete latent variable pretraining method to solve the one-to-many problem of the dialogue system. Meena~\cite{adiwardana2020towards} pretrains a large-scale model with the dialogue data and demonstrates its conversation ability. SC-GPT~\cite{peng2020few} uses a pre-trained language model to convert a dialog act to a natural language response. For the task-oriented dialogue, the large-scale domain-specific dialogue data is inaccessible. The TOD models~\cite{jiang2020convbert,wu2020tod,yu2020score} are usually pretrained on the chit-chat dialogues (Reddit) first and then fine-tuned on the smaller released or synthetic TOD data. Different from the above PLMs, we pretrain the TOD model directly with the large-scale contextual text. We extract relation triples of the contextual text as the grounded ontology-like knowledge and design adaptive self-supervised learning tasks for the end-to-end TOD.

\paragraph{Knowledge-grounded PLMs} Recently, there is an important branch of pre-trained language model to study how to integrate the knowledge into the PLM. ERNIE~\cite{zhang2019ernie} utilizes the external knowledge graph to recognize the type of the mentioned entity. There is a entity type embedding layer as one of input representation. To enhance the knowledge-related representation, they improve the mask mechanism by masking a whole entity directly. Similarly, \citet{rosset2020knowledge} proposes an knowledge-aware language model (KALM), which is decoder-only Transformer-based architecture, like GPT. KALM proposes an entity tokenizer to directly segment popular entities as a single token. Some fields have lots of proprietary information, like medicine, which is emergency to integrate the knowledge. SMedBERT~\cite{zhang2021smedbert} incorporates deep structured semantics knowledge from neighbours of linked-entity. In this paper, we aim to utilize the external tool OpenIE6 to produce lots of TOD-like data to bridge the gap between pre-trained task and end-to-end TOD system. The proposed ontology-like triple recovery task only masks the object values in the extracted triples, rather than randomly masks mentioned entities.

\section{Conclusion \& Future Work}
In this paper, we propose an ontology-aware pretraining method for modeling the end-to-end task-oriented dialogue. The scale of the existing task-oriented dialogue data is far from the need for the pretrained model. Thus, we leverage the external tool OpenIE6 in extracting the ontology-like knowledge of the large-scale contextual texts. To bridge the gap between the pretrained and end-to-end TOD models, we design two adaptive self-supervised learning tasks: ontology-like triple recovery and next-text generation. The pretraining process divides into two phases, where the phase-1 pretrains on the large-scale ontology-aware contextual texts and the phase-2 pretrains on the ontology-aware TOD data. Our proposed OPAL achieves excellent performance on the end-to-end TOD tasks and dialogue state tracking tasks. In the future, we will evaluate the effect of the different ontology-building methods.

\bibliography{tacl2018}
\bibliographystyle{acl_natbib}

\end{document}